# Title of the manuscript

Automated Labeling of German Chest X-Ray Radiology Reports using Deep Learning

# Manuskripttitel

Automatisiertes Labeling Deutscher Röntgenthoraxbefunde durch Deep Learning

## Authors

- Alessandro Wollek, M.Sc.[1,2]
- Philip Haitzer, B.Sc. *[1,2]
- Thomas Sedlmeyr, B.Sc.*[1,2]
- Sardi Hyska, MD. [3]
- Johannes Rueckel, MD. [3,4]
- Bastian O. Sabel, MD. [3]
- Michael Ingrisch, PhD. [3]
- Tobias Lasser, PhD. [1,2]

    * Thomas Sedlmeyr and Philip Haitzer contributed equally to this work.

    [1] Munich Institute of Biomedical Engineering, Technical University of Munich, Garching b. München, Bavaria, Germany

    [2] School of Computation, Information and Technology, Technical University of Munich, Garching b. München, Germany

    [3] Department of Radiology, University Hospital, LMU Munich, Munich, Germany

    [4] Institute of Neuroradiology, University Hospital, LMU Munich, Munich, Germany

## Work originated from

## Corresponding Author

Alessandro Wollek
Phone: +49 89 289 10840
Email: alessandro.wollek@tum.de
Address: Boltzmannstr. 11, Garching b. München, 85748, Bavaria, Germany

## Funding

The research for this article received funding from the German federal ministry of health's program for digital innovations for the improvement of patient-centered care in healthcare [grant agreement no. 2520DAT920].

# Abstract

## Purpose

The aim of this study was to explore the potential of weak supervision in a deep learning-based label prediction model. The goal was to use this model to extract labels from German free-text thoracic radiology reports on chest X-ray images and for training chest X-ray classification models.

## Materials and Methods

The proposed label extraction model for German thoracic radiology reports uses a German BERT encoder as a backbone and classifies a report based on the CheXpert architecture. For investigating the efficient use of manually annotated data, the model was trained using manual annotations, weak rule-based labels, and both. Rule-based labels were extracted from 66,071 retrospectively collected radiology reports from 2017 to 2021 (DS 0), and 1,091 reports from 2020 to 2021 (DS 1) were manually labeled according to the CheXpert classes. Label extraction performance was evaluated with respect to mention extraction, negation detection, and uncertainty detection by measuring F1 scores. The influence of the label extraction method on chest X-ray classification was evaluated on a pneumothorax data set (DS 2) containing 6,434 chest radiographs with associated reports and expert diagnoses of pneumothorax. For this, DenseNet-121 models trained on manual annotations, rule-based and deep learning-based label predictions, and publicly available data were compared.

## Results

The proposed deep learning-based labeler (DL) performed on average considerably stronger than the rule-based labeler (RB) for all three tasks on DS 1 with F1 scores of 0.94 vs. 0.92 for mention extraction, 0.89 vs. 0.51 for negation detection, and 0.61 vs. 0.51 for uncertainty detection, respectively. Pre-training on DS 0 and fine-tuning on DS 1 performed better than only training on either DS 0 or DS 1. Chest X-ray pneumothorax classification results (DS 2) were highest when trained with DL labels with an area under the receiver operating curve (AUC) of 0.939 [95 % CI: 0.925, 0.952] compared to RB labels with an AUC of 0.858 [95 % CI: 0.832, 0.882]. Training with manual labels performed slightly worse than training with DL labels with an AUC of 0.934 [95 % CI: 0.918, 0.949]. In contrast, training with a public data set resulted in an AUC of 0.720 [95 % CI: 0.687, 0.882].

## Conclusion

Our results show that leveraging a rule-based report labeler for weak supervision leads to an improved labeling performance. The pneumothorax classification results demonstrate that our proposed deep learning-based labeler can serve as a substitute for manual labeling requiring only 1,000 manually annotated reports for training.

# Zusammenfassung

## Ziel

Das Ziel dieser Studie war es, das Potenzial der schwachen Supervision in einem auf Deep Learning basierenden Modell zur Extraktion von Labels zu untersuchen. Die Motivation bestand darin, dieses Modell zu verwenden, um Labels aus deutschen Freitext-Thorax-Radiologie-Befunden zu extrahieren und damit Röntgenthorax-Klassifikationsmodelle zu trainieren.

## Material und Methoden

Das vorgeschlagene Modell zur Label-Extraktion für deutsche Thorax-Radiologie-Befunde verwendet einen deutschen BERT-Encoder als Grundlage und klassifiziert einen Befund basierend auf der CheXpert-Architektur. Um den effizienten Einsatz von manuell annotierten Daten zu untersuchen, wurde das Modell mit manuellen Annotationen, regelbasierten Labels und beidem trainiert. Regelbasierte Labels wurden aus 66.071 retrospektiv gesammelten Radiologie-Befunden von 2017 bis 2021 (DS 0) extrahiert, und 1.091 Befunde von 2020 bis 2021 (DS 1) wurden gemäß den CheXpert-Klassen manuell annotiert. Die Leistung der Label-Extraktion wurde anhand der Erfassung von Erwähnungen, der Erkennung von Negationen und der Erkennung von Unsicherheiten anhand von F1-Scores bewertet. Der Einfluss der Label-Extraktionsmethode auf die Röntgenthorax-Klassifikation wurde anhand eines Pneumothorax-Datensatzes (DS 2) mit 6.434 Thoraxaufnahmen und entsprechenden Befunden evaluiert. Hierbei wurden DenseNet-121-Modelle, die mit manuellen Annotationen, regelbasierten und durch Deep Learning-basierten Label-Vorhersagen sowie öffentlich verfügbaren Daten trainiert wurden, verglichen.

## Ergebnisse

Der vorgeschlagene auf Deep Learning basierende Labeler (DL) zeigte im Durchschnitt für alle drei Aufgaben auf DS 1 eine bedeutend bessere Leistung als der regelbasierte Labeler (RB) mit F1-Scores von 0,94 gegenüber 0,92 für die Erwähnungserkennung, 0,89 gegenüber 0,51 für die Negationserkennung und 0,61 gegenüber 0,51 für die Unsicherheitserkennung. Das Vortraining auf DS 0 und das Feintuning auf DS 1 lieferte bessere Ergebnisse als nur das Training auf entweder DS 0 oder DS 1. Die Klassifikationsergebnisse für Pneumothorax auf Röntgenthoraces (DS 2) waren am besten, wenn sie mit DL-Labels trainiert wurden, mit einer Fläche unter der ROC-Kurve (AUC) von 0,939 [95 % CI: 0,925, 0,952], im Vergleich zu RB-Labels mit einer AUC von 0,858 [95 % CI: 0,832, 0,882]. Das Training mit manuellen Labels war etwas schlechter als das Training mit DL-Labels mit einer AUC von 0,934 [95 % CI: 0,918, 0,949]. Das Training mit einem öffentlichen Datensatz führte zu einer AUC von 0,720 [95 % CI: 0,687, 0,882].

## Schlussfolgerung

Unsere Ergebnisse zeigen, dass die Nutzung eines regelbasierten Labelers für schwache Überwachung zu einer verbesserten Labeling-Leistung führt. Die Klassifikationsergebnisse für Pneumothorax zeigen, dass unser vorgeschlagener auf Deep Learning basierender Labeler ein möglicher Ersatz für manuelles Labeling ist und nur 1.000 manuell annotierte Berichte für das Training benötigt.

# Key Points

- The proposed deep learning-based label extraction model for German thoracic radiology reports performs better than the rule-based model.
- Training with limited supervision outperformed training with a small manually labeled data set.
- Using predicted labels for pneumothorax classification from chest radiographs performed equally to using manual annotations.

# Introduction

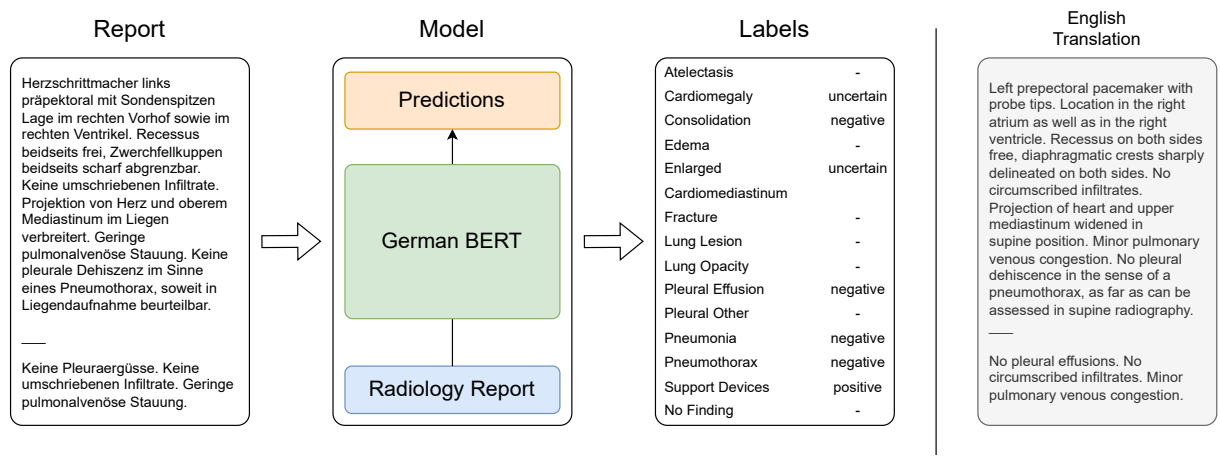

**Figure 1: Automatic label prediction from German thoracic radiology reports. A report is processed by the BERT-based labeler and converted to 14 labels, motivated by the categories in the CheXpert data set. A class is labeled as positive, negative, or uncertain. If the class was not mentioned, it is classified as blank (-).** *Automatische Labelextraktion aus deutschen Thorax-Radiologiebefunden. Ein Befund wird durch den auf BERT basierenden Labeler verarbeitet und in 14 Labels umgewandelt, basierend auf dem CheXpert-Datensatz. Eine Klasse wird als positiv, negativ oder unsicher gekennzeichnet. Wenn die Klasse nicht erwähnt wurde, wird sie als leer (-) klassifiziert.*

Radiologists are in short supply worldwide [1–4], and deep learning models hold promise for addressing this shortage, for example, as part of clinical decision-support systems [5,6]. However, training such models often requires large data sets [7,8] that are expensive and time-consuming to manually label [9,10]. To reduce the amount of time for obtaining labeled data sets, automatic label extraction from radiology reports is a compelling option. Unfortunately, label extraction from radiology reports itself is a challenging task, for example, due to missing annotated data [11].

Recent developments in the natural language processing (NLP) domain have proposed models that generate dense word vector representations [12–15] which have shown to be effective in training deep learning models for a wide range of tasks such as translation [16] or named entity recognition [17]. Similar to the computer vision domain, these language models can be pre-trained on a general, large corpus and then fine-tuned on a target corpus that might be otherwise too small for training [18].

In the medical domain, language models have been successfully applied to extract labels from unstructured radiology reports. Smit et al. improved upon their rule-based labeler for English radiology reports by using a BERT [14] language model as a backbone [19]. Similarly, Nowak et al. investigated the use of BERT for German radiology reports [10]. They compared a rule-based labeler to a deep learning model, trained with 18,000 manually annotated reports, rule-based extracted labels, and a combination of both.

In this work, we explore the potential of weak supervision of a deep learning-based label prediction model, using a rule-based labeler [20]. The general label extraction pipeline is illustrated in **Figure 1**. Our proposed label extraction model takes a German free-text thoracic radiology report and extracts the corresponding labels. In contrast to Nowak et al., we focus on the classes of the CheXpert data set [21], allowing for comparison with previous works and pooling of data sets for future studies. More importantly, we study the effect of extracted labels on downstream image classification training. Our study builds upon previous work that used rule-based strategies to extract labels [20]. We conduct extensive experiments on a data set of internal radiology reports and our results demonstrate the effectiveness of our approach.

Our contributions are:

- We propose a deep learning-based label extraction model for German thoracic radiology reports.
- We demonstrate that our labeler outperforms a rule-based label extraction model with respect to label extraction and utility for downstream applications.
- We show that a pneumothorax classifier trained with automatically extracted labels performs equivalently to a model trained on manual annotations.

Our code is publicly available at https://gitlab.lrz.de/IP/german-chexbert.

## Materials and Methods

### Data Collection

| Split | Data Set 0 (DS 0) | Data Set 1 (DS 1) | Data Set 2 (DS 2) |
|---|---:|---:|---:|
| Training | 60,071 | 203 — 810 | 4,507 |
| Validation | 1,000 | 51 — 203 | 660 |
| Test | 5,000 | 78 | 1,267 |
| Total | 66,071 | 1,091 | 6,434 |
| Annotations | automatic | report, manual | CXR + report, manual |

**Table 1: Data sets used in this study. Data set 0 (DS 0) was labeled with a rule-based labeler [20], data set 1 (DS 1) was manually annotated solely based on radiological reports, and data set 2 (DS 2) was labeled based on the chest radiographs (CXR) and radiological reports. To measure the importance of available data, training and validation splits of DS 1 were split into quarters.** *Datensätze, die in dieser Studie verwendet wurden. Datensatz 0 (DS 0) wurde mit einem regelbasierten Labeler [20] gelabelt, Datensatz 1 (DS 1) wurde ausschließlich auf der Grundlage radiologischer Befunde manuell annotiert, und Datensatz 2 (DS 2) wurde anhand von Röntgenthoraces (CXR) und radiologischen Befunden annotiert. Um die Bedeutung der verfügbaren Daten zu messen, wurden Trainings- und Validierungsdaten von DS 1 in Viertel aufgeteilt.*

| Dataset | Data Set 1 (DS 1) | | | | | | Data Set 2 (DS 2) | | | | | |
|---|---|---|---|---|---|---|---|---|---|---|---|---|
| Class | P | U | N | P | U | N | P | N | P | N | P | N |
| Atelectasis | 220 | 54 | 2 | 12 | 13 | 1 | - | - | - | - | - | - |
| Cardiomegaly | 184 | 368 | 266 | 16 | 25 | 25 | - | - | - | - | - | - |
| Consolidation | 205 | 45 | 627 | 23 | 6 | 41 | - | - | - | - | - | - |
| Edema | 297 | 9 | 521 | 24 | 5 | 34 | - | - | - | - | - | - |
| Enlarged Cardiom. | 223 | 295 | 305 | 22 | 19 | 26 | - | - | - | - | - | - |
| Fracture | 63 | 3 | 79 | 9 | 2 | 8 | - | - | - | - | - | - |
| Lung Lesion | 44 | 7 | 8 | 5 | 5 | 5 | - | - | - | - | - | - |
| Lung Opacity | 278 | 41 | 565 | 28 | 6 | 35 | - | - | - | - | - | - |
| No Finding | 1 | 0 | 0 | 1 | 0 | 0 | - | - | - | - | - | - |
| Pleural Effusion | 455 | 45 | 451 | 29 | 11 | 32 | - | - | - | - | - | - |
| Pleural Other | 57 | 16 | 1 | 7 | 5 | 0 | - | - | - | - | - | - |
| Pneumonia | 52 | 173 | 649 | 5 | 16 | 45 | - | - | - | - | - | - |
| Pneumothorax | 83 | 7 | 871 | 5 | 5 | 66 | 1122 | 3385 | 204 | 456 | 326 | 941 |
| Support Devices | 590 | 1 | 107 | 43 | 1 | 12 | - | - | - | - | - | - |

**Table 2: Label distributions of manually annotated data sets used in this study. Data set 1 class annotations were labeled based on free text reports [20]. Data set 2 class annotations were based on reports and radiographs [22]. Enlarged Cardiom. = Enlarged Cardiomediastinum, P = Positive, U = Uncertain, N = Negative.** *Die Klassenverteilung der verwendeten manuell annotierten Datensätze. Bei Datensatz 1 erfolgte die Klassifizierung basierend auf Befunden [20]. Bei Datensatz 2 wurden die Klassifizierungen anhand von Befunden und Röntgenaufnahmen vorgenommen [22]. Enlarged Cardiom. = Vergrößertes Cardiomediastinum, P = Positiv, U = Unsicher, N = Negativ.*

Data splits and annotation methods of all data sets used throughout this study are reported in **Table 1**. We retrospectively identified 66,071 thoracic radiology reports from 2017 to 2021 in our institutional PACS (DS 0). Additionally, we used 1,091 thoracic radiology reports from 2020-2021 that were manually annotated by a first-year radiology resident from Klinikum der Ludwig-Maximilians-Universität München in a previous study [20]. In the following, we refer to the manually annotated reports as data set 1 (DS 1).

The training and test set label distributions of DS 1 are reported in **Table 2**. Since annotated "no finding" reports describe normal appearing chest radiographs, there are no negative or uncertain annotations available for this class.

To increase the number of training samples, we favored test samples with multiple non-blank annotations. We selected 78 of the 1,091 reports of DS 1 for testing. Our selection process ensured that each class was mentioned by at least five reports, whenever available. In cases where the entire data set contained less than five samples for a specific class, half of the samples were designated for testing. None of the 78 DS 1 reports used for testing were part of DS 0.

To further test our model we utilized another internal data set consisting of 6,434 chest radiographs with according reports [22]. We refer to this data set in the following as data set 2 (DS 2). This data set, in contrast to DS 1, contains only binary pneumothorax annotations. However, the annotations are based on both report and chest radiograph providing a higher label quality. In the data set, 1,568 samples have been labeled as pneumothorax.

## Architecture

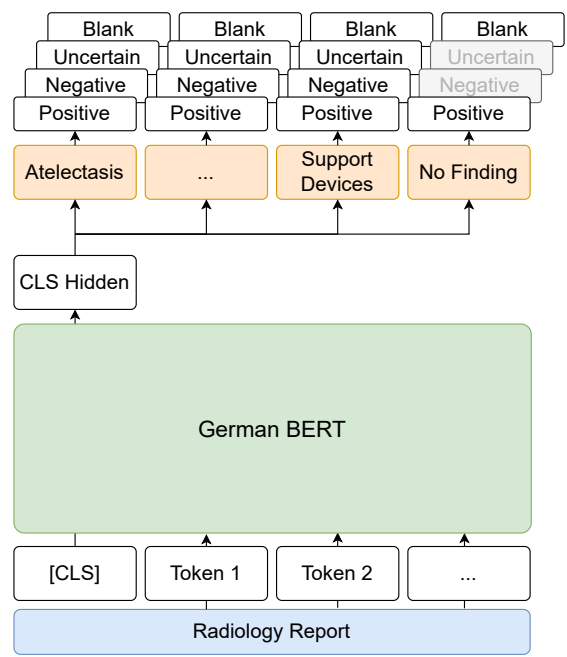

**Figure 2: Deep learning-based German radiology report labeler. The model extracts CheXpert labels from free-text radiology reports.** *Deep Learning-basierter deutscher Radiologiebefund-Labeler. Das Modell extrahiert die CheXpert-Labels aus unstrukturierten Befundtexten.*

Following Smit et al. [19], we used a pre-trained BERT [14] model as a backbone for our label extraction model. The objective of the model is to predict the fourteen CheXpert [21] labels: atelectasis, cardiomegaly, consolidation, edema, enlarged cardiomediastinum, fracture, lung lesion, lung opacity, pleural effusion, pleural other, pneumonia, pneumothorax, support devices, and "no finding" given a German radiology report.

The architecture is illustrated in Figure 2. The model receives the report as an input and assigns one of the classes: blank, positive, negative, or uncertain to each of the 13 categories, mirroring a manual annotation. The blank classification represents no mention of the class in the report. For the special case "no finding", which corresponds a normal report, the labeler must predict only blank or positive.

We modified the BERT architecture by using 14 linear heads, as illustrated in Figure 2. Each head is dedicated to capture one of the 14 labels. For transfer learning, we use the pre-trained "bert-base-german-cased" BERT model[1] trained on German texts, such as the German Wikipedia corpus, with a sequence length of 512 tokens.

To predict the classes of the 14 findings, the radiology reports were tokenized first. Of all tokenized reports, a single report in the training data, and none in the test data consisted of more than 512 tokens. The overflowing report consisted of 579 tokens and described multiple images. We considered only the first 512 tokens of this report. After tokenization, the reports were processed by the model. Subsequently, the hidden state of the class (CLS) token from the final layer was used as the input for each of the 14 linear heads, predicting the class of each finding via a softmax.

The model was fine-tuned using cross-entropy loss, AdamW [23] optimization with default parameters ($\beta_1 = 0.9$, $\beta_2 = 0.999$), a learning rate of 2e-5, and a batch size of 8. The individual cross-entropy losses for the 14 observations were aggregated before calculating the final loss. To monitor model performance, we periodically evaluated the model on the validation set and saved the best checkpoint across all 14 observations.

## Label Extraction (DS 1)

We evaluated our deep learning-based labeler on the three tasks proposed by the original CheXpert data set: mention extraction, negation detection, and uncertainty detection. Following the original CheXpert experimental setup, findings labeled as "blank" were considered as negative for the mention extraction task and the other classes ("positive", "negative", or "uncertain") as positive. Regarding negation detection, only the "negative" classification was considered positive, and for uncertainty detection, only the "uncertain" class was considered positive.

To assess the importance of manually and automatically extracted annotations, we designed three experiments: training only with manually annotated reports (supervised), DS 1, automatically extracted labels (weakly supervised), DS 0, and all available data (hybrid), DS 0 + DS 1.

### Supervised Approach

As a baseline, we trained the model solely on manually annotated reports (DS 1). To assess the importance of the number of annotated reports, we split the training data into quarters and trained on increasingly larger fractions, as reported in **Table 3**.

| Run | Training | Validation | Test |
|---|---|---|---|
| 25 % | 203 | 51 | 78 |
| 50 % | 406 | 101 | 78 |
| 75 % | 608 | 152 | 78 |
| 100 % | 810 | 203 | 78 |

---

[1] https://huggingface.co/bert-base-german-cased

**Table 3: Training, validation, and test splits for the different training runs on DS 1.** *Training, Validierung und Testaufteilungen für die verschiedenen Experimente auf DS 1.*

## Weakly Supervised Approach

We investigated the benefit of weak labels on label extraction performance. The labels were created using the rule-based model proposed in a previous study [20]. For validation, we randomly sampled 1,000 reports, and for internal testing 5,000 reports, without patient overlap, from the total 66,071 reports of DS 0 (see **Table 1**). We used the remaining 60,071 reports for training. For final testing, we used the manually labeled test reports of DS 1.

## Hybrid Approach

To leverage all available data, we fine-tuned the weakly supervised model on the manually annotated reports (DS 1). Again, we trained the model on increasing fractions of DS 1, as reported in **Table 3**.

## Pneumothorax Classification (DS 2)

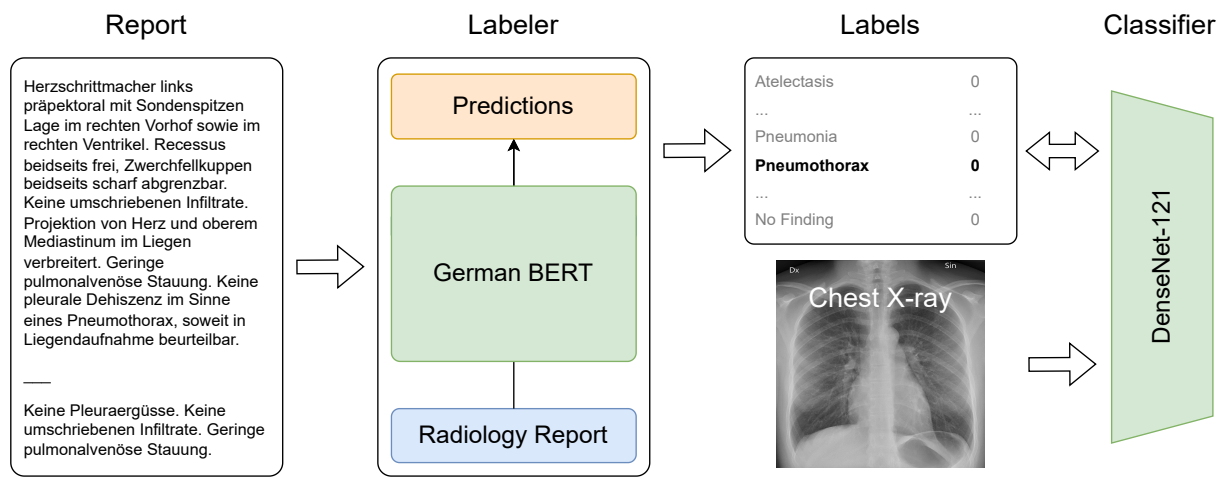

**Figure 3: Pneumothorax classification model trained with automatically extracted annotations.** *Pneumothorax Klassifikationsmodell, trainiert mit automatisch extrahierten Annotationen.*

To address the limitation of the small test data set of DS 1, we tested the labeler on the larger data set 2. Since the data set contains only binary pneumothorax annotations, we considered uncertain predictions as positive, and blank predictions as negative.

Furthermore, as the goal of label extraction is the training of image classification models, we trained a DenseNet-121 [24] to predict the presence of a pneumothorax on chest radiographs based on manual and extracted labels.

Our pneumothorax classification pipeline utilized a DenseNet-121 pre-trained on ImageNet as a backbone. We replaced the final fully-connected layer with a one-dimensional version for fine-tuning on DS 2. The final softmax activation was replaced by a sigmoid. Training involved 10 epochs with AdamW with default parameters ($\beta_1$ = 0.9, $\beta_2$ = 0.999), a learning rate of 0.003, and batch size of 32. We selected the checkpoint for the final model based on the validation area under the receiver operating characteristic curve (AUC). All images were resized to 224x224 pixels and normalized using the ImageNet mean and standard deviation. Data augmentation involved ten-crop, i.e., taking five crops of the regular and flipped image. The complete pipeline for the deep learning-based experimental setup is illustrated in **Figure 3**.

We assessed the effect of labeling method on pneumothorax classification performance on DS 2 by comparing fine-tuning using radiologists' annotations [22], rule-based [20] or deep learning-based extracted labels, with a DenseNet-121 fine-tuned on the chest X-ray 14 data set [25] (CheXnet [26]).

## Statistical Evaluation

For all three experimental settings on DS 1 we measured mean F1 scores for the three tasks of mention extraction, negation detection, and uncertainty detection by comparing model predictions with manually annotated test reports.

Label extraction performance on DS 2 was measured using sensitivity and specificity. To simplify the comparison with DS 1, we applied the same metrics. We measured pneumothorax classification performance by analyzing receiver operating characteristics (ROC) and AUC. As our research involves numerous comparisons and is purely explorative, we abstained from reporting P values and instead presented 95 % confidence intervals, which were calculated using 10,000-fold resampling via non-parametric bootstrap methodology at the level of the image or report. Due to space limitations, 95 % confidence intervals for the F1 scores were not included.

All statistical analyses were performed using NumPy version 1.24.2 and Scikit-Learn version 1.2.2.

# Results

## Label Extraction (DS 1)

| Data Set 1 | Mention Extraction | | | Negation Detection | | | Uncertainty Detection | | |
|---|---|---|---|---|---|---|---|---|---|
| Run | S | WS | H | S | WS | H | S | WS | H |
| 25 % | 0.87 | - | 0.93 | 0.69 | - | 0.88 | 0.59 | - | 0.56 |
| 50 % | 0.82 | - | **0.94** | 0.72 | - | 0.87 | 0.43 | - | **0.64** |
| 75 % | 0.87 | - | **0.94** | 0.81 | - | **0.89** | 0.61 | - | 0.54 |
| 100 % | 0.84 | 0.91 | **0.94** | 0.83 | 0.82 | **0.89** | 0.57 | 0.52 | 0.61 |

**Table 4: Comparison of mean F1 scores for mention extraction, negation detection, and uncertainty detection. Supervised (S) and hybrid (H) models were trained on different fractions of data set 1 training data. Weakly supervised (WS) and hybrid models were (pre-) trained on data set 0.** *Vergleich der durchschnittlichen F1-Scores für das Extrahieren von Erwähnungen, das Erkennen von Verneinungen und das Erkennen von Unsicherheit. Überwachte (S) und hybride (H) Modelle wurden mit verschiedenen Anteilen der Trainingsdaten von Datensatz 1 trainiert. Schwach überwachte (WS) und hybride Modelle wurden mit Datensatz 0 (vor-) trainiert.*

### Supervised Approach

The results obtained when trained solely on increasing fractions of DS 1, are reported in **Table 4**. Mention extraction F1 scores ranged from 82 % to 84 %. Negation detection F1 scores improved from 69 % to 83 % when increasing the amount of training data. Mean uncertainty detection F1 scores ranged from 43 % to 61 %.

### Weakly Supervised Approach

When trained only with reports labeled by the rule-based labeler (DS 0), the model achieved a mean F1 score of 91 % for mention extraction, 82 % for negation detection, and 52 % for uncertainty detection (see **Table 4**). Note that although the model was trained on DS 0, the reported test results were measured on DS 1.

### Hybrid Approach

The effect of pre-training with automatically labeled reports first and then fine-tuning on varying amounts of manually annotated data is reported in **Table 4**. Mention extraction F1 scores ranged from 93 % to 94 %, with a slightly higher score obtained with more training data. Similarly, negation detection F1 scores improved slightly when using more manually annotated training data, F1 scores ranged from 0.88 % to 89 %. Mean uncertainty detection F1 scores ranged from 56 % to 64 %.

| Data Set 1 | Mention Extraction | | Negation Detection | | Uncertainty Detection | |
|---|---|---|---|---|---|---|
| Findings | RB | DL | RB | DL | RB | DL |
| Atelectasis | **0.982** | 0.963 | **1.0** | N/A | **0.769** | 0.700 |
| Cardiomegaly | 0.667 | **0.955** | 0.649 | **0.898** | 0.571 | **0.809** |
| Consolidation | 0.950 | **0.979** | 0.746 | **0.911** | **0.400** | **0.400** |
| Edema | **0.992** | 0.985 | 0.939 | **0.955** | **0.600** | N/A |
| Enlarged Cardiomediastinum | 0.820 | **0.933** | **0.800** | 0.776 | 0.500 | **0.821** |
| Fracture | **0.900** | **0.900** | 0.545 | **0.857** | N/A | N/A |
| Lung Lesion | 0.857 | **0.938** | 0.889 | 0.889 | 0.182 | **0.714** |
| Lung Opacity | 0.936 | **0.952** | 0.667 | **0.853** | **0.316** | N/A |
| No Finding | **0.025** | N/A | - | - | - | - |
| Pleural Effusion | 0.973 | **0.974** | 0.954 | **0.955** | 0.556 | **0.706** |
| Pleural Other | **0.857** | 0.737 | N/A | N/A | **0.500** | 0.333 |
| Pneumonia | 0.922 | **0.964** | 0.892 | **0.966** | 0.600 | **0.688** |
| Pneumothorax | 0.987 | **0.994** | **0.964** | 0.950 | **0.571** | 0.333 |

| | | | | | |
|---|---|---|---|---|---|
| Support Devices | **0.972** | 0.962 | **0.800** | 0.762 | N/A | N/A |
| Mean | 0.91 | **0.94** | 0.82 | **0.89** | 0.51 | **0.61** |

**Table 5: Rule-based (RB) and deep learning-based (DL) label extraction F1 scores for the three evaluation tasks: mention extraction, negation detection, and uncertainty detection for each finding. Labels were extracted from DS 1 and compared to manual annotations. N/A results could not be calculated due to insufficient data. Higher values are highlighted in bold.**

*Regelbasierte (RB) und Deep Learning-basierte (DL) Label-Extraktions-F1-Scores für die drei Evaluationstasks: Extraktion von Erwähnungen, Negationserkennung und Unsicherheitserkennung für jede Klasse. Die Labels wurden aus DS 1 extrahiert und mit manuellen Annotationen verglichen. N/A-Ergebnisse konnten aufgrund unzureichender Daten nicht berechnet werden. Höhere Werte sind fett hervorgehoben.*

| | Data Set 1 | | | |
|---|---|---|---|---|
| | Sensitivity | | Specificity | |
| **Findings** | **RB** | **DL** | **RB** | **DL** |
| Atelectasis | **0.960** [0.867-1.000] | 0.920 [0.800-1.000] | **1.000** [1.000-1.000] | 0.981 [0.939-1.000] |
| Cardiomegaly | 0.561 [0.405-0.714] | **0.927** [0.838-1.000] | **0.892** [0.781-0.976] | 0.838 [0.710-0.946] |
| Consolidation | **0.966** [0.886-1.000] | 0.897 [0.769-1.000] | 0.735 [0.605-0.852] | **0.857** [0.750-0.945] |
| Edema | **0.966** [0.886-1.000] | 0.966 [0.885-1.000] | 0.918 [0.833-0.981] | **0.959** [0.896-1.000] |
| Enlarged Cardiomediastinum | 0.659 [0.512-0.800] | **0.854** [0.737-0.953] | **0.784** [0.645-0.909] | 0.730 [0.579-0.867] |
| Fracture | **0.909** [0.700-1.000] | **0.909** [0.700-1.000] | 0.940 [0.877-0.986] | **0.985** [0.952-1.000] |
| Lung Lesion | **0.900** [0.667-1.000] | **0.900** [0.667-1.000] | 0.956 [0.900-1.000] | **1.000** [1.000-1.000] |
| Lung Opacity | **0.971** [0.903-1.000] | 0.882 [0.765-0.974] | 0.636 [0.489-0.773] | **0.818** [0.698-0.927] |
| No Finding | 0.000 [0.000-0.000] | 0.000 [0.000-0.000] | 0.922 [0.857-0.974] | **1.000** [1.000-1.000] |
| Pleural Effusion | **0.925** [0.833-1.000] | **0.925** [0.833-1.000] | **0.974** [0.914-1.000] | **0.974** [0.914-1.000] |
| Pleural Other | **0.750** [0.500-1.000] | 0.583 [0.286-0.875] | **1.000** [1.000-1.000] | **1.000** [1.000-1.000] |
| Pneumonia | 0.857 [0.696-1.000] | **0.952** [0.842-1.000] | **0.982** [0.943-1.000] | **0.982** [0.943-1.000] |
| Pneumothorax | **0.600** [0.273-0.900] | 0.400 [0.100-0.727] | 0.971 [0.925-1.000] | **0.985** [0.954-1.000] |
| Support Devices | **0.932** [0.850-1.000] | 0.909 [0.818-0.979] | 0.941 [0.850-1.000] | **0.971** [0.903-1.000] |
| Mean | 0.782 | **0.787** | 0.904 | **0.934** |

|  | Data Set 2 | | | |
| --- | --- | --- | --- | --- |
|  | Sensitivity | | Specificity | |
| Findings | RB | DL | RB | DL |
| Pneumothorax | **0.997** [0.994, 0.999] | 0.972 [0.963-0.979] | 0.991 [0.988, 0.994] | **0.995** [0.993-0.997] |

**Table 6: Sensitivity and specificity of the extracted labels compared to the reference annotations on DS 1 and DS 2 with corresponding 95 % confidence intervals. To create binary labels, uncertain labels/annotations were considered positive, blank negative. The deep learning model was first pre-trained on weak labels and then fine-tuned on 100 % of the manually annotated training data. RB = rule-based labeler, DL = deep learning-based labeler (ours). Higher values are highlighted in bold.** *Sensitivität und Spezifität der extrahierten Labels im Vergleich zu den Referenzannotationen für DS 1 und DS 2 mit entsprechenden 95 % Konfidenzintervallen. Um binäre Labels zu erstellen, wurden unsichere Labels/Annotationen als positiv und leere als negativ betrachtet. Das Deep-Learning-Modell wurde zunächst mit schwachen Labels vortrainiert und anschließend mit 100 % der manuell annotierten Trainingsdaten gefinetuned. RB = regelbasierte Labeler, DL = Deep-Learning-basierter Labeler (unserer). Höhere Werte sind fett hervorgehoben.*

Rule-based and deep learning-based label extraction results for all three evaluation tasks are compared in **Table 5**. The deep learning-based labeler was pre-trained with labels extracted by the rule-based labeler and fine-tuned on 100 % of the manually annotated training data. Across all three tasks, the deep learning model performed better. For mention extraction, our proposed labeler had a mean F1 score of 94 % compared to 91 % of the rule-based labeler. For negation and uncertainty detection, the improvement of using a deep learning-based labeler compared to a rule-based model was even greater, with 89 % vs. 82 % mean F1 score for negation detection, and 61 % vs. 51 % mean F1 score for uncertainty detection.

To simplify comparison of labeling results on DS 1 with the labeling results on DS 2, we additionally measured sensitivity and specificity by considering uncertain labels as positive and blank labels as negative. The results are reported in **Table 6**. On average, the deep learning-based labeler achieved a higher sensitivity compared to the rule-based approach with 0.787 vs. 0.782 and a higher specificity with 0.934 vs. 0.904.

## Pneumothorax Label Extraction (DS 2)

The comparison of rule-based and deep learning-based labeler for pneumothorax annotation on DS 2 is presented in **Table 6**. The rule-based labeler had a higher sensitivity compared to the deep learning-based model with 0.997 [95 % CI: 0.994, 0.999] vs. 0.972 [95 % CI: 0.963, 0.979]. In contrast, the deep learning-based labeler had a higher specificity with 0.995 [95 % CI: 0.993, 0.997] vs. 0.991 [95 % CI: 0.988, 0.994].

# Pneumothorax Classification (DS 2)

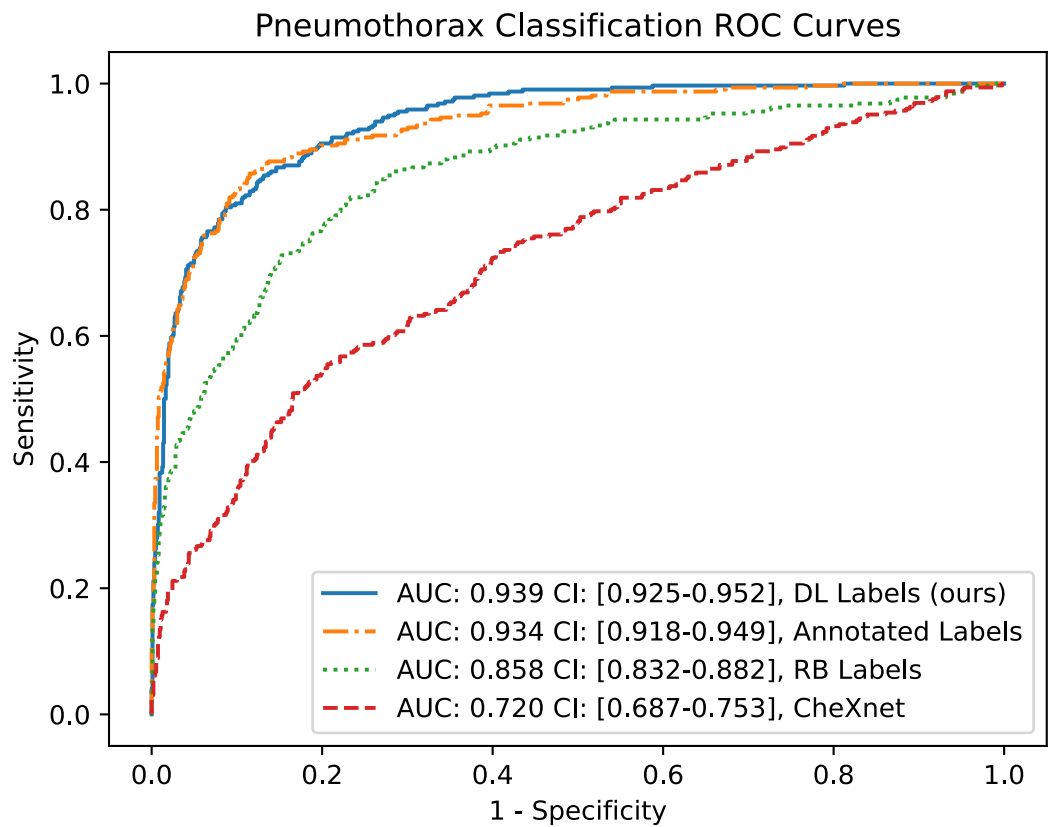

**Figure 4: Pneumothorax classification receiver operating characteristic (ROC) curves and areas under the ROC curve (AUC) with corresponding 95 % confidence intervals. All models, except the CheXnet baseline, were trained on DS 2 with manual expert annotations (Annotated Labels), extracted with a rule-based (RB Labels), or our proposed deep learning-based labeler (DL Labels). The CheXnet model was trained on the chest X-ray 14 data set.** *Pneumothorax-Klassifikation Receiver Operating Characteristic (ROC) Kurven und Flächen unter der ROC-Kurve (AUC) mit entsprechenden 95 % Konfidenzintervallen. Alle Modelle außer dem CheXnet-Baseline-Modell wurden entweder mit manuellen Expertenannotationen (Annotated Labels) aus DS 2, mit Labels extrahiert durch den regelbasierten Ansatz (RB Labels) oder von unserem vorgeschlagenen Labeler basierend auf Deep Learning (DL Labels) trainiert. Das CheXnet-Modell wurde mit dem Chest X-ray 14 Datensatz trainiert.*

To assess the performance of our proposed label extraction algorithm we trained a pneumothorax classifier on chest radiographs with labels generated by different methods. The classification ROC curves and AUC values with corresponding 95 % confidence intervals are shown in **Figure 4**.

The baseline CheXnet model trained on the chest X-ray 14 data set achieved lowest perfromance with an AUC of 0.720 [95 % CI: 0.687, 0.882], followed by the model trained on DS 2 with labels extracted from the rule-based model with an AUC of 0.858 [95 % CI: 0.832, 0.882]. When trained on labels created by radiologists inspecting both image and report the model achieved an AUC score of

0.934 [95 % CI: 0.918, 0.949]. Highest AUC values were obtained when trained with labels extracted by our proposed deep learning-based model with 0.939 AUC [95 % CI: 0.925, 0.952].

# Discussion

In this work, we proposed a deep learning-based chest radiology report label extraction model. The best performing model was pre-trained on reports labeled by a rule-based labeler and fine-tuned on only a thousand manually labeled reports. On average, it substantially outperformed the rule-based model, on all three tasks, with a mean mention extraction F1 score of 94 % vs. 91 %, negation detection F1 score of 89 % vs. 82 %, and uncertainty F1 score of 61 % vs. 51 % (see **Table 5**). These results show that the improvements of employing a deep learning-based compared to a rule-based label extraction model transfer from English to German radiology reports [21].

The pneumothorax classification results provide further evidence of the improvements of our proposed deep learning-based labeler compared to the rule-based labeler. Not only did the AUC increase from 0.858 to 0.939, but it also surpassed the model trained on the DS 2 labels that were annotated by radiologists based on inspecting the image and report. These results suggest that training with labels extracted by the deep learning-based labeler is an alternative to the time-consuming manual labeling.

Similar to Nowak et al. [10], the deep learning-based model outperformed the rule-based model on German reports. Apart from using different data sets, a direct comparison is difficult, as they considered both uncertain and negative mentions as negative labels. Furthermore, their rule-based labeler achieved only an average classification F1 score of 75.1 %, compared to their deep learning-model with 95.5 %. Therefore, they observed that pre-training with weakly labeled reports did not improve the deep learning model performance. In contrast, our rule based-labeler served as a strong baseline (see **Table 5**). Consequently, pre-training with such weak supervision improved the performance compared to only training on manually annotated data alone. For example, mean mention extraction F1 score improved from 84 % to 94 % when using all data. Furthermore, our model was trained on only approximately 1,000 manually labeled reports, compared to a total of 18,000 used by Nowak et al. [10]. While they showed that increasing the amount of manually annotated training data improved mean F1 scores from 70.9 % to 95.5 % when increasing training data from 500 to 14,580 samples, annotating all 18,000 samples took 197 hours. However, based on their results, we assume that increasing the number of manually annotated samples could further improve our model.

Our study has several limitations. First, due to the limited number of available manually annotated reports, most data were used for training. To compensate for this, we tested the model on a larger data set (DS 2). A future study with more manually annotated data could both improve model performance and reduce the variance of test scores. Another limitation is that the labels of DS 1 were created by a single radiologist, possibly introducing label biases or errors made due to annotation fatigue.

In conclusion, we demonstrated a significant improvement in German radiology report labeling using our proposed deep learning-based labeler, achieving a new state-of-the-art on these data sets. Our results provide evidence of the benefits of employing a deep learning-based model, even in scenarios with sparse data, and the use of the rule-based labeler as a tool for weak supervision.

## Clinical Relevance

One of the main motivations of employing deep learning models in clinical decision support systems is to reduce the effects of the worldwide shortage of radiologists. However, the data to train such models must be annotated by radiologists. Our presented labeler drastically reduces the required amount of manually annotated reports and even outperformed the pneumothorax classification model trained with labels created by radiologists.

## Conflict of Interest

The authors declare that they have no conflict of interest.